\documentclass{egpubl}

\SpecialIssuePaper         

\CGFccby

\usepackage[T1]{fontenc}
\usepackage{dfadobe}  

\biberVersion
\BibtexOrBiblatex
\usepackage[backend=biber,bibstyle=EG,citestyle=alphabetic,backref=true]{biblatex} 
\electronicVersion
\PrintedOrElectronic
\ifpdf \usepackage[pdftex]{graphicx} \pdfcompresslevel=9
\else \usepackage[dvips]{graphicx} \fi

\usepackage{egweblnk}

\title[Navigating Perplexity]%
      {Navigating Perplexity:\\ A linear relationship with the data set size in t-SNE embeddings}

\author[M.~Skrodzki, N.\,F.~Chaves-de-Plaza, T.~H\"ollt, E.~Eisemann \& K.~Hildebrandt]
{\parbox{\textwidth}{\centering 
    M.~Skrodzki\thanks{Corresponding author: \href{mailto:mail@ms-math-computer.science}{mail@ms-math-computer.science}}$^{1}$\orcid{0000-0002-8126-0511},
    N.\,F.~Chaves-de-Plaza$^{1}$\orcid{0000-0003-4971-3151},
    T.~H\"ollt$^{1}$\orcid{0000-0001-8125-1650},
    E.~Eisemann$^{1}$\orcid{0000-0003-4153-065X},
    and K.~Hildebrandt$^{1}$\orcid{0000-0002-9196-3923}
        }
        \\
{\parbox{\textwidth}{\centering $^1$Computer Graphics and Visualization, TU Delft, the Netherlands
       }
}
}

\graphicspath{{images/}}

\usepackage{amsmath}
\usepackage{amsfonts}

\DeclareMathOperator{\KL}{KL}

\DeclareMathOperator{\per}{Perp}

\usepackage{algorithm}
\usepackage{algpseudocode}

\algnewcommand\algorithmicforeach{\textbf{for each}}
\algdef{S}[FOR]{ForEach}[1]{\algorithmicforeach\ #1\ \algorithmicdo}

\newtheorem{proposition}{Proposition}

\begin{document}


\maketitle
\begin{abstract}
    Widely used pipelines for analyzing high-dimensional data utilize two-dimensional visualizations. 
    These are created, for instance, via t-distributed stochastic neighbor embedding (t-SNE). 
    A crucial element of the t-SNE embedding procedure is the \emph{perplexity} hyperparameter.
    That is because the embedding structure varies when perplexity is changed.
    A suitable perplexity choice depends on the data set and the intended usage for the embedding.
    Therefore, perplexity is often chosen based on heuristics, intuition, and prior experience.
    
    This paper uncovers a linear relationship between perplexity and the data set size.
    Namely, we show that embeddings remain structurally consistent across data set samples when perplexity is adjusted accordingly.
    Qualitative and quantitative experimental results support these findings.
    This informs the visualization process, guiding the user when choosing a perplexity value.
    Finally, we outline several applications for the visualization of high-dimensional data via t-SNE based on this linear relationship.


\end{abstract}  
\section{Introduction}

Modern visualization workflows frequently use dimensionality reduction methods for visually inspecting high-dimensional data.
A prominent technique that has made a resounding impact on the visualization of such data is \emph{t-distributed Stochastic Neighbor Embedding} (t-SNE)~\cite{vandermaaten2008visualizing}.
It excels at revealing local structures and is particularly effective in visualizing clusters and relationships within complex data sets.

While t-SNE has a broad impact in various application domains supported by available software libraries, conceptual and computational challenges remain that have the potential to substantially improve t-SNE's utility for applications. 
One such challenge is the choice of suitable parameter settings, specifically for the \emph{perplexity} hyperparameter. 
Variations of the perplexity fundamentally affect the resulting embeddings' structures~\cite{wattenberg2016how}.
Yet, as the original publication suggested perplexity values between~5 and~50, this stuck as the default.
That is despite the fact that other works have formulated heuristics for the choice of perplexity that rather depend on the size of the data set to be embedded and thus lie well outside of the originally proposed interval~\cite{belkina2019automated,kobak2019art}.

Despite the availability of these heuristics, finding a suitable perplexity value for a given data set is often tedious.
That is because the perplexity depends on the structure of the specific data set and the embedding's intended usage.
Hence, operators of visualization pipelines have to resort to intuition, which is built on prior experience.
This is important since, for large data sets, creating embeddings with various perplexity values quickly becomes computationally expensive.
Therefore, obtaining different t-SNE embeddings with various perplexities is practically infeasible.

This paper shows that embeddings remain structurally consistent across data set samples when the perplexity is adjusted according to a linear relationship with the sample's size.
Thereby, we preserve an embedding's key visual characteristics across differently sized data set samples.
The linear relationship is established theoretically and validated via quantitative and qualitative experiments.

Our results lend themselves well to practical applications. 
For instance, searching for a suitable perplexity value can be accelerated by first determining a perplexity value on samples and then computing the corresponding perplexity value for the full data set.
This allows users to quickly gauge the effect of a certain perplexity choice without having to compute the full embedding.
Furthermore, we augment a sampling-based embedding approach that aims to build and maintain coarse-level structures when prolongating it to the full data set size.
Finally, the linear relationship can be used to estimate embeddings for large perplexity values that would otherwise be too computationally intensive to be computed.
For instance, users can sample a data set to an extent that can be run on their machine with an intended perplexity value, even if the full data set cannot be embedded with a large perplexity.

In summary, in this article, we:
\begin{itemize}
    \item \ldots establish a theoretical background for a linear relation between the size of a data set and the perplexity, see Section~\ref{sec:PerplexityScalesLinearlyWithPointSetSize}.
    \item \ldots quantitatively and qualitatively validate the linear relationship for several data sets, see Section~\ref{sec:ExperimentalValidation}.
    \item \ldots present several consequences of the linear relationship for practical visualization workflows, see Section~\ref{sec:ApplicationsOfTheLinearRelationship}.
\end{itemize}


\section{Background: T-SNE dimensionality reduction}

Methods for dimensionality reduction can be classified according to whether their embedding is obtained linearly or non-linearly and whether they aim to preserve local or global distances.
Here, we focus on t-SNE~\cite{vandermaaten2008visualizing}, a non-linear, locally preserving method.
Other methods in this class include LLE (Locally Linear Embedding)~\cite{roweis2000nonlinear}, LE (Laplacian Eigenmaps)~\cite{belkin2001laplacian}, LAMP (Local Affine Multidimensional Projection)~\cite{joia2011local}, and UMAP~\cite{becht2019dimensionality}.
We refer to a recent survey for the advantages and disadvantages of the respective classes and methods~\cite{xia2021revisiting}.
The survey states that non-linear embedding techniques, such as t-SNE, ``preserve local neighborhood[s] in [the] D[imensionality] R[eduction] processes''.
Furthermore, they find that t-SNE ``perform[s] the best in cluster identification and membership identification.''
This motivates our focus on t-SNE.

T-SNE creates a low-dimensional data embedding while aiming at preserving local neighborhoods of the high-dimensional data points~\cite{vandermaaten2008visualizing}.
This is achieved by interpreting the high-dimensional input~${D={\{\mathbf{x}_1,\ldots,\mathbf{x}_n\}\subseteq\mathbb{R}^d}}$ as conditional probabilities by
\begin{align}
\label{equ:HighDimensionalProbability}
    p_{j|i} = \frac{\exp\left(-\left\|\mathbf{x}_i-\mathbf{x}_j\right\|^2/2\sigma_i\right)}{\sum_{k\neq i}\exp\left(-\left\|\mathbf{x}_i-\mathbf{x}_k\right\|^2/2\sigma_i^2\right)}, \quad 
    p_{ij} = \frac{p_{j|i}+p_{i|j}}{2},
\end{align}
where~$p_{i|i}=0$ and~$\sigma_i$ is the variance of the Gaussian centered on point~$\mathbf{x}_i$.
In practice, the bandwidth~$\sigma_i$ is chosen such that the perplexity of the probability distribution~${P_i=\{p_{j|i} \mid j\in[n]\}}$ equals a user-prescribed perplexity value~$\per$ by
\begin{align}
\label{eq:PerplexityEntropy}
    \per = 2^{-\sum_j p_{j|i}\log(p_{j|i})}.
\end{align}
On the low-dimensional embedding~${\{\mathbf{y}_1,\ldots,\mathbf{y}_n\}\subseteq\mathbb{R}^{d'}}$, a corresponding probability distribution is given by
\begin{align}
\label{equ:LowDimensionalProbability}
    q_{ij}=\frac{\left(1+\left\|\mathbf{y}_i-\mathbf{y}_j\right\|^2\right)^{-1}}{\sum_{k\neq \ell}\left(1+\left\|\mathbf{y}_k - \mathbf{y}_\ell\right\|^2\right)^{-1}}.
\end{align}
To compute the positions~$\mathbf{y}_i$ of the low-dimensional embedding, t-SNE starts with an initial embedding obtained by principal component analysis (PCA)~\cite{kobak2021initialization} and then alters the embedding by gradient-descent optimization of the Kullback-Leibler divergence between the high- and the low-dimensional probability distribution, which is given by
\begin{align}
\label{equ:KLDivergence}
    C = \KL(P||Q) = \sum_i\sum_j p_{ij}\log \frac{p_{ij}}{q_{ij}},
\end{align}
with the gradient
\begin{align}
\label{equ:KLGradient}
    \frac{\partial C}{\partial \mathbf{y}_i} = 4\sum_{j\neq i}(p_{ij}-q_{ij})\left(1+\left\|\mathbf{y}_i-\mathbf{y}_j\right\|^2\right)^{-1}(\mathbf{y}_i-\mathbf{y}_j).
\end{align}
Usually, in the first several hundred iterations of gradient descent, the high-dimensional probabilities are exaggerated by multiplying~$P$ with a factor.
A commonly used factor for the \emph{early exaggeration} is~$12$, as established in a comparative study~\cite{belkina2019automated}.
The same study suggests an initial learning rate of~$n/12$, which is then augmented by momentum and gain~\cite{jacobs1988increased,vandermaaten2008visualizing} to increase the convergence of the optimization.

The naive implementation of t-SNE has a run time of~$\mathcal{O}(n^2)$ per iteration.
Therefore, several acceleration structures have been proposed.
A wide-spread acceleration is given by the Barnes-Hut (BH) tree structure~\cite{vandermaaten2014accelerating}.
The main idea is that the gradient can be split into terms representing the attractive and repulsive forces.
The former is fast to evaluate, given a sparse matrix~$P$ of high-dimensional probabilities.
The latter can then be approximated by aggregating points to node centers of a quad tree.
A similar approach utilizes the Fast-Fourier Transform to approximate the repulsive forces~\cite{linderman2019fast}.
Here, the domain is discretized and the forces are first computed on the discrete nodes.
Then, the forces for each embedding point are approximated from the surrounding nodes.
Other acceleration structures that utilize parallelization and GPU functionality are available~\cite{Ulyanov2016,chen2019robust,pezzotti2020gpgpu,vanderuit2022efficient}.

In addition to t-SNE, UMAP provides another popular method of creating embeddings~\cite{becht2019dimensionality,mcinnes2020UMAP}.
However, due to it using stochastic gradient descent, the relationship between UMAP's \emph{neighborhood size} hyperparameter and the obtained embeddings is less clear than for t-SNE.
Furthermore, recent work suggests that UMAP-type embeddings can be interpreted as a part of a spectrum of embeddings, attainable via t-SNE~\cite{boehm2022attraction-repulsion,damrich2023from}, which makes the two approaches closely related.
Thus, the following discussion will be restricted to t-SNE.






\section{Related work on the perplexity hyperparameter}
\label{sec:ThePerplexityHyperparameter}

The arguably most important hyperparameter when using t-SNE to embed a data set is the perplexity~$\per$.
It steers embedding properties that depend on the specific use case, for instance, how spread out a cluster should be or how many clusters are created~\cite{wattenberg2016how}. 
In this sense, there is no one-size-fits-all choice for perplexity.
Previous work observed that ``the optimal perplexity parameter increased as the total number of data points increased``~\cite{ding2018interpretable} in the context of single-cell transcriptomics.
Therefore, their approach chooses~${\per=10n/512}$, where~$n$ is the number of data points to be embedded.
Similarly, when choosing a balance between nearest-neighbor and global structure preservation, the heuristic~${\per=n/100}$ has been proposed~\cite{kobak2019art}.
Applying these heuristics leads to reliable and replicable results since these fix a ratio between the input data size and the perplexity.
Note that they differ drastically from the recommendations of basic t-SNE, which chose values between~${\per=5}$ and~${\per=50}$~\cite{vandermaaten2008visualizing}.
Several works set out to better understand and estimate suitable perplexity values.

One possibility is to select perplexity in a post hoc fashion.
First, compute embeddings for ten perplexity values~${\per=5}$ to~${\per=50}$.
Then, measure Kruskal's stress and the Spearman rank correlation to determine which perplexity value gives the most faithful embedding~\cite{waagen2021tSNE}.
While this produces reliable results and ensures selecting a suitable perplexity value within the investigated range, it is an extremely costly approach as all embeddings must be computed.
Other work acknowledges that different perplexity values lead to subjectively better or worse embeddings.
In a similar post hoc fashion, they embed a data set for perplexity~${\per\in\{15,50,100,200,500}\}$ and choose the best-suited embedding via visual inspection~\cite{sharma2023optimization}.
Particularly for large-scale data sets, this post hoc approach quickly becomes infeasible due to the long computation times.

Another approach automatically finds a suitable perplexity by solving a separate optimization problem~\cite{cao2017automatic}.
Here, the final cost function value from Equation~(\ref{equ:KLDivergence}) is related to the term~$\log(n)\per/n$.
The automatically chosen~$\per$ values are validated in a user study.
Similarly to the previous approach, it is possible to alter the formulation of the high-dimensional probabilities in Equation~(\ref{equ:HighDimensionalProbability}).
Namely, introducing multi-scale Gaussian similarities~\cite{lee2015multi} allows sampling of the bandwidths in a data-driven fashion~\cite{debodt2018perplexity,crecchi2020perplexity}.
Yet another related approach starts with a low perplexity value that likely leads to undefined values in the probability matrix~$P$ due to isolated vertices in the high-dimensional neighborhood graph.
Either perplexity or the bandwidths~$\sigma_i$ is then iteratively increased until a suitable probability matrix without undefined values is found~\cite{xiao2023optimizing}.
While these approaches require little to no extra computation aside from the t-SNE optimization, they lead to a single perplexity and embedding by design.
This contrasts previous findings highlighting how different perplexity values can reveal different structures and aspects of a data set~\cite{wattenberg2016how,sharma2023optimization}.
While they remove the need for a user-given perplexity parameter, they also remove the freedom to explore different embeddings.

Such freedom is provided by the interactive approach of approximated t-SNE (A-tSNE)~\cite{pezzotti2016approximated}.
The authors approximate the distances~$\mathcal{D}$ using approximated K-Nearest Neighborhood (KNN) queries.
This allows the user to investigate the first embeddings of the data almost immediately after it has been loaded.
Throughout the embedding process, the user can decide on local refinements and steer the approximation level during the analysis.
The authors use a fixed perplexity value of~${\per=30}$ in their experiments.
Similarly to approximating the neighborhoods in A-TSNE, the dependence on the specific metric used in computing the probability distribution, Equation~(\ref{equ:LowDimensionalProbability}), can be alleviated by taking multiple metric maps into account~\cite{shen2017visualization}. 
This multiple maps (mm-TSNE) variant uses a constant~$\sigma$ for the variances in the conditional probabilities, Equation~(\ref{equ:HighDimensionalProbability}), which completely removes user control via perplexity.
Therefore, our investigations improve the output of both these t-SNE variations.


\section{Embeddings remain structurally similar when scaling perplexity linearly with sample sizes}
\label{sec:PerplexityScalesLinearlyWithPointSetSize}

In this section, we will present the main result of our paper.
It will be experimentally validated in Section~\ref{sec:ExperimentalValidation}.
To proceed, first recall that the choice of the perplexity value has a significant effect on the structure of the embedding~\cite{wattenberg2016how}.
Also, the heuristics~\cite{ding2018interpretable,kobak2019art} discussed above lead to a certain structural type of embedding, when other structures might also be insightful.
Such different types are obtained by choosing a different perplexity when keeping the number of points constant or changing the number of points while keeping the perplexity constant.
However, when varying the perplexity or the data set size, the user needs to understand the effects of this variation.
To help this understanding, we provide the following statement:

\begin{proposition}
\label{pro:MainResult}
    Given a low-dimensional embedding~$E$ of some high-dimensional input data~$D$, obtained via t-SNE with a perplexity value~$\per$.
    Then, an embedding~$E'$ of a $\rho$-subset~${D'\subset D}$ will preserve key visual characteristics of~$E$ if it is obtained with a perplexity value~${\per' = \rho\cdot\per}$, where ${\rho=\frac{|D'|}{|D|}}$.
\end{proposition}

Before we move on to substantiate this statement, note that it only applies to a certain regime.
The perplexity values~$\per$ and~$\per'$ determine the bandwidths~$\sigma_i$ used to compute the high-dimensional probabilities from Equation~(\ref{equ:HighDimensionalProbability}).
No relationships between the high-dimensional points can be established if these bandwidths are too narrow.
Thus, there is an implicit lower bound on~$\per'$ and therefore on~$\rho$.

\begin{figure}
    \centering
    \def\svgwidth{1.\linewidth}
\begingroup%
  \makeatletter%
  \providecommand\color[2][]{%
    \errmessage{(Inkscape) Color is used for the text in Inkscape, but the package 'color.sty' is not loaded}%
    \renewcommand\color[2][]{}%
  }%
  \providecommand\transparent[1]{%
    \errmessage{(Inkscape) Transparency is used (non-zero) for the text in Inkscape, but the package 'transparent.sty' is not loaded}%
    \renewcommand\transparent[1]{}%
  }%
  \providecommand\rotatebox[2]{#2}%
  \newcommand*\fsize{\dimexpr\f@size pt\relax}%
  \newcommand*\lineheight[1]{\fontsize{\fsize}{#1\fsize}\selectfont}%
  \ifx\svgwidth\undefined%
    \setlength{\unitlength}{447.54503145bp}%
    \ifx\svgscale\undefined%
      \relax%
    \else%
      \setlength{\unitlength}{\unitlength * \real{\svgscale}}%
    \fi%
  \else%
    \setlength{\unitlength}{\svgwidth}%
  \fi%
  \global\let\svgwidth\undefined%
  \global\let\svgscale\undefined%
  \makeatother%
  \begin{picture}(1,0.32053721)%
    \lineheight{1}%
    \setlength\tabcolsep{0pt}%
    \put(0,0){\includegraphics[width=\unitlength,page=1]{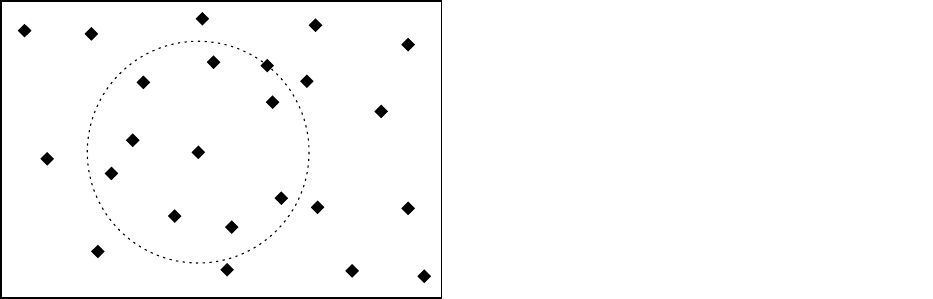}}%
    \put(0.19858268,0.14416304){\color[rgb]{0,0,0}\makebox(0,0)[lt]{\lineheight{1.25}\smash{\begin{tabular}[t]{l}$\mathbf{x}_i$\end{tabular}}}}%
    \put(0.17762735,0.19416647){\color[rgb]{0,0,0}\makebox(0,0)[lt]{\lineheight{1.25}\smash{\begin{tabular}[t]{l}$d_i$\end{tabular}}}}%
    \put(0,0){\includegraphics[width=\unitlength,page=2]{sampling_rationale.pdf}}%
    \put(0.72406032,0.14416302){\color[rgb]{0,0,0}\makebox(0,0)[lt]{\lineheight{1.25}\smash{\begin{tabular}[t]{l}$\mathbf{x}_i$\end{tabular}}}}%
    \put(0.70310509,0.19416647){\color[rgb]{0,0,0}\makebox(0,0)[lt]{\lineheight{1.25}\smash{\begin{tabular}[t]{l}$d_i$\end{tabular}}}}%
    \put(0,0){\includegraphics[width=\unitlength,page=3]{sampling_rationale.pdf}}%
  \end{picture}%
\endgroup%

    \caption{
        Left: On the full data set, when creating a neighborhood of~${k=9}$ points around a point~$\mathbf{x}_i$, this neighborhood lies in a ball of radius~$d_i$. 
        Right: Taking a~${\rho=0.5}$ sample (sampled points remain black, non-sampled points are faded), we expect to have~${\rho\cdot k=4.5}$ sampled points within the same sized neighborhood.
    }
    \label{fig:samplingRationale}
\end{figure}

Consider the following scheme to formalize the above statement.
If on the full data level, we aim to work with neighborhoods of~$k$ points, this fixes a certain distance~$d_i$ around each point~${\mathbf{x}_i\in D}$ within which the~$k$ neighbors lie, respectively.
We then consider a $\rho$-sample~$D'$ of the data~$D$.
For each point~${\mathbf{x}_i\in D'}$, we expect to find~${\rho\cdot k}$ many points from the sample~$D'$ within distance~$d_i$ of~$\mathbf{x}_i$, see Figure~\ref{fig:samplingRationale}.
Therefore, if we search for~${\rho\cdot k}$ neighbors of a point~$\mathbf{x}_i\in D'$ in a $\rho$-sampling~$D'$ we expect the furthest neighbor to lie at about distance~$d_i$ from the center point.
This distance~$d_i$ is important as it, together with the bandwidth~$\sigma_i$, determines the high-dimensional probabilities from Equation~(\ref{equ:HighDimensionalProbability}).
However, the value of~$\sigma_i$ is chosen via a binary search based on the user-given perplexity~\cite{vandermaaten2008visualizing}.
Hence, when reducing or increasing the number of points in a data set, we observe an interplay of the neighborhood sizes and the perplexity.

\subsection{Sparse high-dimensional neighborhoods}

To make the above reasoning more explicit, we will first consider an approximation of t-SNE, where a neighborhood size~$k$ is given or derived as~${k=\min\{n, 3\cdot\per\}}$~\cite{vandermaaten2014accelerating}.
That is, instead of taking all~$n$ points into account when evaluating Equation~(\ref{eq:PerplexityEntropy}), only the~$k$ nearest neighbors of point~$\mathbf{x}_i$ are used.
Then, for a given high-dimensional data point~$\mathbf{x}_i\in D$, its $k$-nearest neighbors~${\{\mathbf{x}'_1,\ldots,\mathbf{x}'_k\}}$ define a geometric neighborhood ball of radius~${d(\mathbf{x}_i,\mathbf{x}'_k)}$, assuming that~$\mathbf{x}'_k$ is the farthest neighbor among the $k$-nearest and denoting the high-dimensional distance by~$d(.,.)$.
Within this geometric neighborhood, the bandwidths~$\sigma_i$ from Equation~(\ref{equ:HighDimensionalProbability}) are chosen to be in-line with the user-provided perplexity value~$\per$ according to Equation~(\ref{eq:PerplexityEntropy}).

When restricting to a uniformly chosen subset~${D'\subseteq D}$, picking each point with probability~${\rho\in(0,1]}$, the expected number of points to lie within the~${d(\mathbf{x}_i,\mathbf{x}'_k)}$-ball around~$\mathbf{x}_i$ is 
\begin{align*}
    &\mathbb{E}\left(\left|D'\cap\{\mathbf{x}\in\mathbb{R}^d\mid d(\mathbf{x},\mathbf{x}_i \leq d(\mathbf{x}_i,\mathbf{x}'_k)\}\right|\right)\\
     = &\mathbb{E}\left(\sum_{i=1}^{k} \mathbf{1}_{\mathbf{x}'_i\in D'}\right)
     = \sum_{i=1}^{k} \mathbb{E}\left(\mathbf{1}_{\mathbf{x}'_i\in D'}\right)
     = \sum_{i=1}^{k} \rho = k \rho,
\end{align*}
with~$\mathbf{1_.}$ the characteristic function of the random variable.
Hence, to provide a point~$\mathbf{x}_i$ with the same geometric information, that is, having all neighbors within the~${d(\mathbf{x}_i,\mathbf{x}'_k)}$-ball with high probability, we have to choose a neighborhood size of~$\rho\cdot k$ or a perplexity of~${\rho\cdot\per}$ respectively.

\subsection{General high-dimensional neighborhoods}

The above reasoning applies whenever the neighborhood size is directly related to the perplexity, as suggested by previous work~\cite{vandermaaten2014accelerating} and implemented, for instance, in \href{https://opentsne.readthedocs.io/en/latest/examples/01_simple_usage/01_simple_usage.html}{OpenTSNE}.
However, in the basic formulation of t-SNE~\cite{vandermaaten2008visualizing}, independent of the given perplexity value~$\per$, all points~$\mathbf{x}_j$ go into the computation of Equation~(\ref{eq:PerplexityEntropy}), not just the~$k$ nearest.
That makes the analysis more involved than in the finite case above.

To get a structurally similar embedding when computing the embedding of a $\rho$-sample of a data set, we want to choose a perplexity~$\per'$ for the sample embedding such that the points' bandwidths~$\sigma_i$ remain the same under the sampling.
The choice of~$\per'$ should only depend on~$\rho$, not on the data set~$D$, nor the specific sample drawn.
Let~${D'=\{\mathbf{x}'_1,\ldots,\mathbf{x}'_{\lceil n\rho\rceil}\}}$ be any sample of~$D$.
Then we want to choose a perplexity~$\per'$ such that the bandwidths~$\sigma'_i$ provided from~$\per'$ by binary search in Equation~(\ref{eq:PerplexityEntropy}) is as similar as possible to the corresponding~$\sigma_i$ on the full data set.

To determine the perplexity~$\per'$ suitable for a specific point~$\mathbf{x}_i$ that was chosen as point~$\mathbf{x}'_j$ of the sample~$D'$, consider~${i\in [n]}$ fixed and take into account the following set of random variables~$X_j$ given by:
\begin{align*}
    X_j=\begin{cases}\exp\left(-d(\mathbf{x}_i,\mathbf{x}_j)^2/2\sigma_i^2\right) & \rho\\0 & 1-\rho\end{cases},
\end{align*}
which are Bernoulli variables of value~${c_j=\exp\left(-d(\mathbf{x}_i,\mathbf{x}_j)^2/2\sigma_i^2\right)}$ with uniform probability~$\rho$ and value~$0$ otherwise.
This is a direct consequence of our choice to sample the data uniformly randomly.
Note that this uses the established bandwidth~$\sigma_i$ from the full data.
These give rise to a new set of random variables
\begin{align*}
    X_{j|i} = \frac{X_j}{\sum_{k\neq i}X_k}.
\end{align*}
Then, we can describe the sought-for perplexity~$\per'$ for the sample point~$\mathbf{x}_i$ as the expectancy of Equation~(\ref{eq:PerplexityEntropy}) with respect to the random variables~$X_{j|i}$ representing the sample:
\begin{align}
\label{equ:PerplexityCoarse}
    \per'(\mathbf{x}_i) &= \mathbb{E}\left(2^{-\sum_{j}X_{j|i}\log(X_{j|i})}\right)
\end{align}
Due to the coupling of the random variables within the exponent, this expression cannot be analyzed analytically.
However, it can be evaluated via an experimental Monte Carlo approach, see Section~\ref{sec:QuantitativeValidationOfTheLinearRelationship}.


\section{Experimental Validation}
\label{sec:ExperimentalValidation}

The previous section proved the linear relationship between perplexity and data set size for the discrete case.
Here, we will experimentally validate the relationship also for the basic formulation of t-SNE~\cite{vandermaaten2008visualizing}, where all points are taken into account when evaluating Equation~(\ref{equ:HighDimensionalProbability}).
We will provide quantitative and qualitative results to support our theoretical statements.

\subsection{Quantitative validation of the linear relationship}
\label{sec:QuantitativeValidationOfTheLinearRelationship}

We use several data sets~$D$ and proceed with a fixed perplexity~${\per=30}$ for these experiments as follows.
We iterate through the sampling rates~${\rho\in\{0.1,\ldots,0.9\}}$.
Then, we draw three uniformly random samples from each of the full data sets for each sampling rate.
For a given sample~$D'$ and a point~${\mathbf{x}_i\in D'}$, we compute the bandwidth~$\sigma_i$ of this point via binary search according to perplexity~$\per$ on the full data set.
That is, we determine~$\sigma_i$ based on Equation~(\ref{equ:HighDimensionalProbability}) taking all points from~$D$ into account.
Then, based on~$\sigma_i$, we compute the conditional probabilities
\begin{align}
    \label{equ:SampledHighDimensionalProbabilities}
    p'_{j|i} = \frac{\exp\left(-\left\|\mathbf{x}_i-\mathbf{x}_j\right\|^2/2\sigma_i\right)}{\sum_{k\neq i}\exp\left(-\left\|\mathbf{x}_i-\mathbf{x}_k\right\|^2/2\sigma_i^2\right)}, \quad \mathbf{x}_j,\mathbf{x}_k\in D',
\end{align}
which provides a probability distribution on the sample~$D'$.
Based on this probability distribution, we compute a sample perplexity
\begin{align}
    \label{equ:SampledPerplexity}
    \per'(\mathbf{x}_i) = 2^{-\sum_j p'_{j|i} \log(p'_{j|i})}
\end{align}
for the point~$\mathbf{x}_i$, which corresponds to the perplexity that would have been necessary for point~$\mathbf{x}_i$ to have bandwidth~$\sigma_i$ on the sample~$D'$.
This evaluates Equation~(\ref{equ:PerplexityCoarse}) for the sampled probabilities.

\begin{algorithm}
    \caption{Monte Carlo Evaluation of Perplexity Scaling}
    \label{alg:MonteCarloEvaluation}
    \begin{algorithmic}
        \Require a $\rho$-sample~$D'\subset D$, perplexity~$\per=30$
        \ForEach {$\mathbf{x}_i\in D'$}  
            \State Find~$\sigma_i$ by Equation~(\ref{equ:HighDimensionalProbability}) with all points from~$D$ and~$\per=30$.
            \State Compute~$p'_{j|i}$ by Equation~(\ref{equ:SampledHighDimensionalProbabilities}) with points from~$D'$ and~$\sigma_i$ from above.
            \State Compute~$\per'(\mathbf{x}_i)$ by Equation~(\ref{equ:SampledPerplexity}) with~$p'_{j|i}$ from above. 
        \EndFor
        \Return Collection of~$\per'(\mathbf{x}_i)$.
    \end{algorithmic}
\end{algorithm}

Note that because we run the experiment on a random sample, we obtain a different perplexity~$\per'(\mathbf{x}_i)$ for every point~$\mathbf{x}_i$ in the sample~$D'$.
However, based on the reasoning from Section~\ref{sec:PerplexityScalesLinearlyWithPointSetSize}, we expect a linear behavior across the different sampling rates.
Statistics across all the sample's points and the three different samples drawn allow estimating the best perplexity~$\per'$ that overall leads to as many sample bandwidths corresponding to the bandwidths observed on the full data set.

\begin{figure}
    \centering
    \includegraphics[width=1.\linewidth]{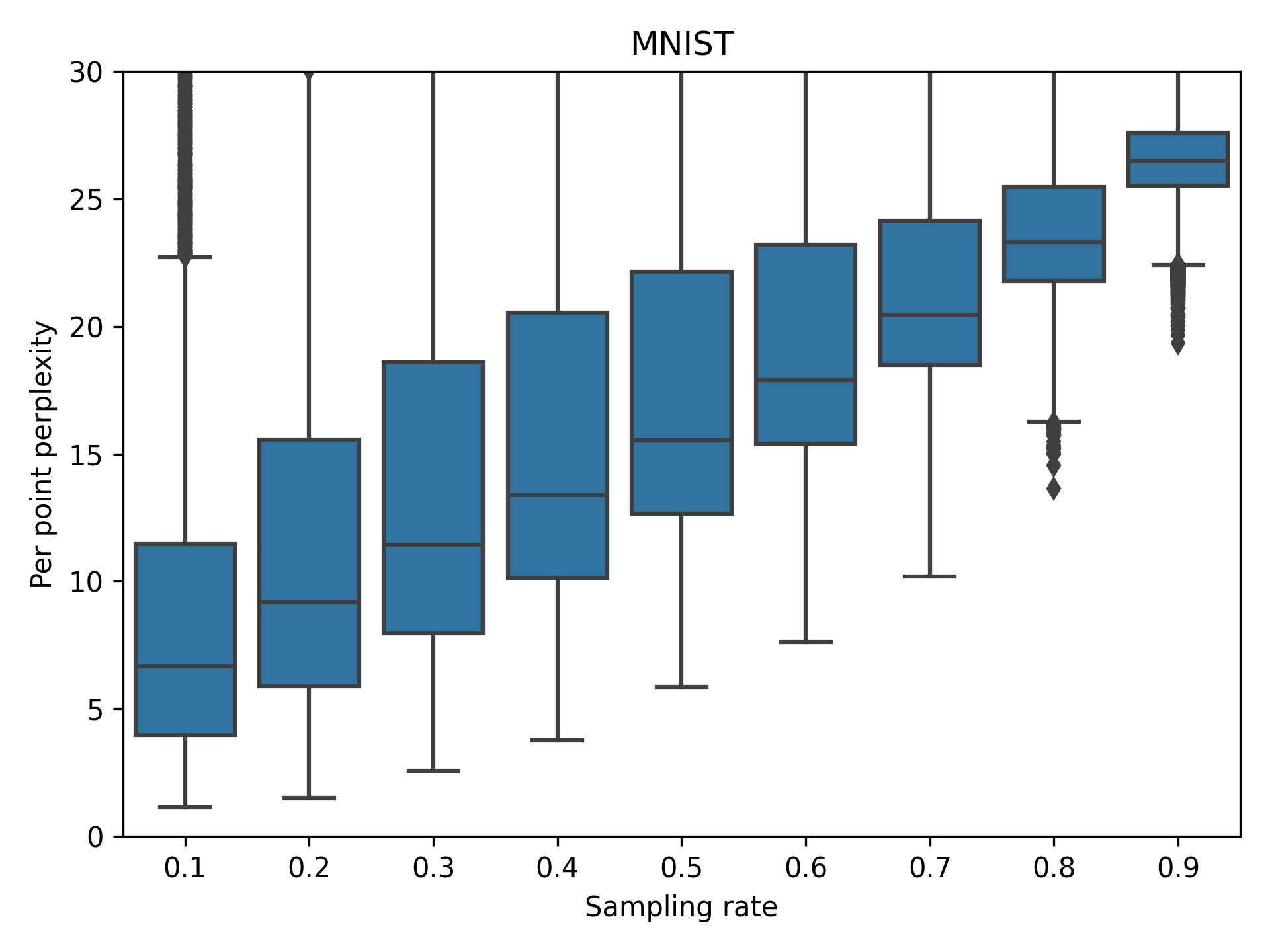}
    \includegraphics[width=1.\linewidth]{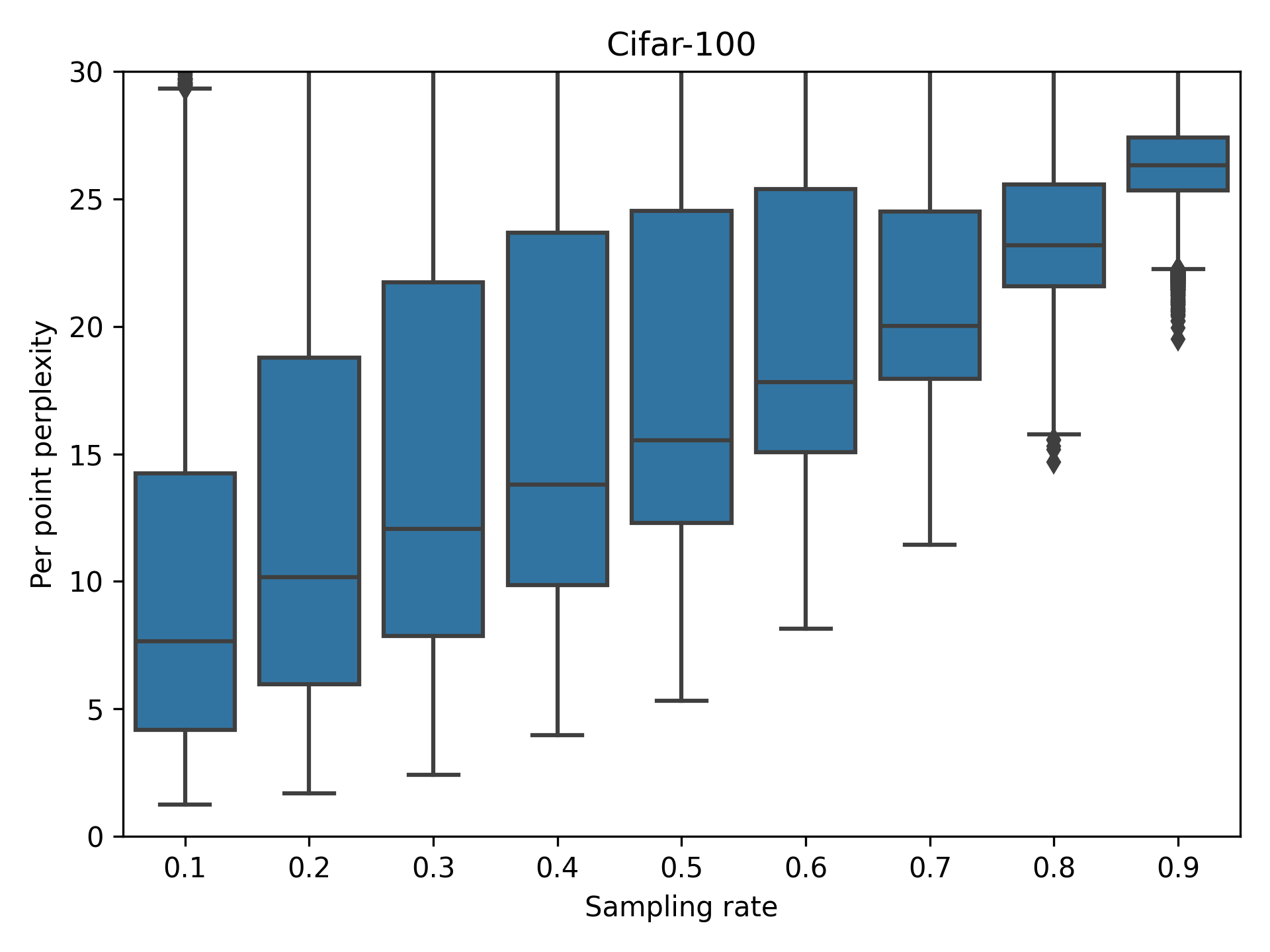}
    \includegraphics[width=1.\linewidth]{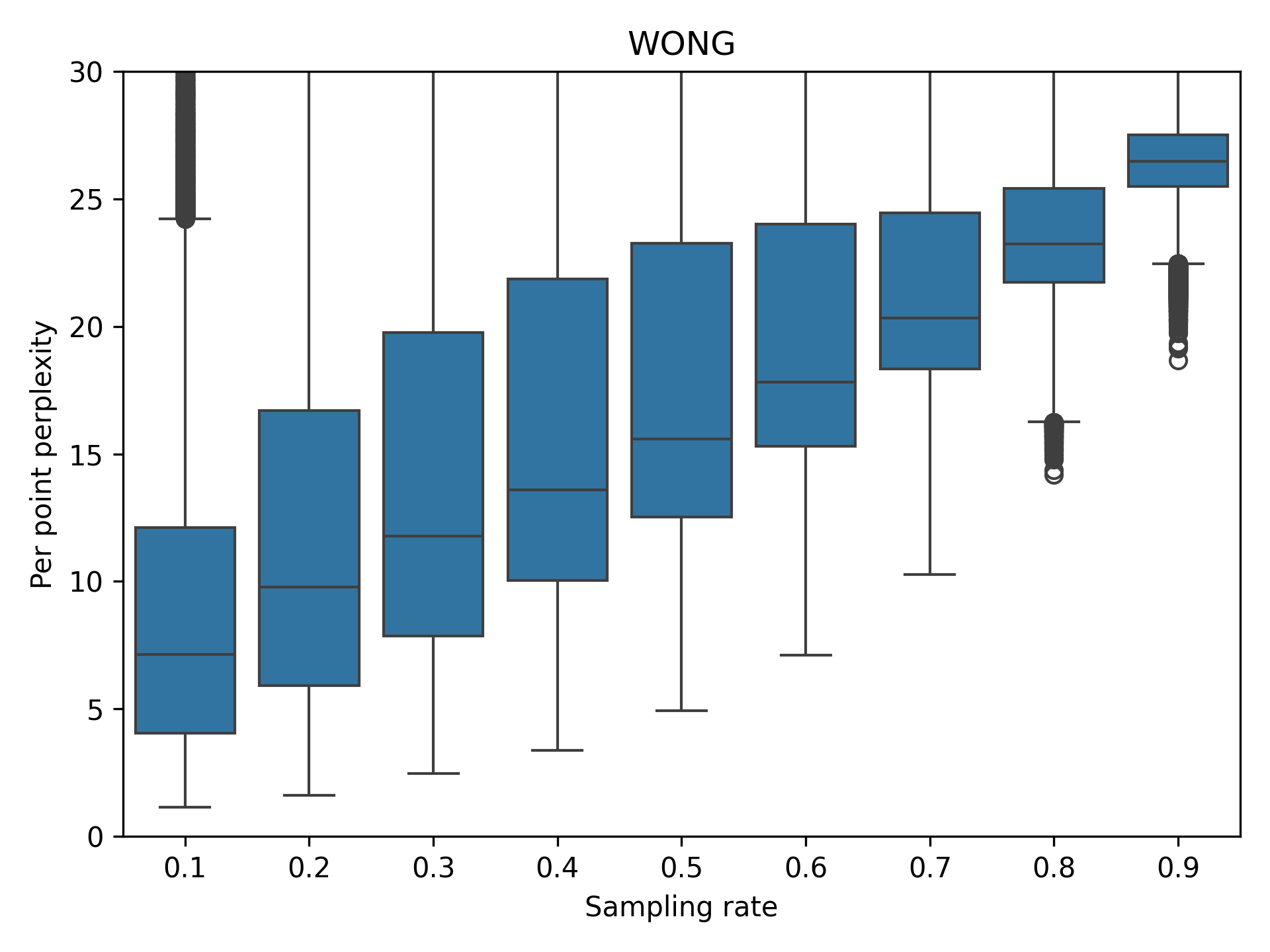}
    \caption{
        Results of our Monte Carlo experiments, Algorithm~\ref{alg:MonteCarloEvaluation}, to estimate the per-point perplexity across different samples of various sizes, drawn from the MNIST, CIFAR-100, and WONG data.
    }
    \label{fig:MonteCarloPerplexity}
\end{figure}

For the evaluation, we turn to the MNIST~$(n=70,000)$,  CIFAR-100~$(n=60,000)$, and WONG~$(n=327,457)$ data set.
MNIST contains of~70,000 hand-written images of the ciphers~0 to~9, while CIFAR-100 consists of~60,000 32x32~color images in~10 classes.
The WONG data contains mass cytometry measurements of T Cell lineages.
Figure~\ref{fig:MonteCarloPerplexity} includes the results of our Monte Carlo experiment on these data sets.
While the specific perplexities vary across a sampling rate, the respective median values clearly show a linear trend.
This trend is anchored at the chosen perplexity~${\per=30}$ which would be attained for the sampling rate~${\rho=1.0}$ for both data sets, which converges in the upper-right corner of the plots.

The data's structure calls for differences in the chosen perplexity values across the different data sets.
However, the linear relationship can be observed clearly.
This validates the linear relationship between perplexity and data set size for the basic t-SNE case taking all points into account~\cite{vandermaaten2008visualizing}.

\subsection{Qualitative validation of the linear relationship}
\label{sec:QualitativeValidationOfLinearRelationship}

We will now provide a qualitative validation of the linear relationship established in Section~\ref{sec:PerplexityScalesLinearlyWithPointSetSize}.
For this, we turn to the MNIST~${(n=70,000)}$, C.Elegans~${(n=89,701)}$, and WONG~${(n=327,457)}$ data sets.
The C.Elegans data set is an experimentally obtained gene expression atlas. 
We compute different variations of perplexity and sampling rates on each data set to investigate their interplay.

First, we compute an initial embedding for each full data set via PCA~\cite{kobak2021initialization}.
Then, we sample the data uniformly to~${\rho\in\{0.7, 0.4, 0.1\}}$.
We obtain an initial embedding for each sample by sampling the PCA embedding of the full data set.
Note that these samples are drawn in a nested fashion, that is, the~${\rho=0.4}$ sample is drawn from the points in the~${\rho=0.7}$ sample and the the~${\rho=0.1}$ sample is drawn from the points in the~${\rho=0.4}$ sample.
This is done to eliminate any unwanted side effects of the random sampling on the visual quality of the embeddings.

Second, we fix four perplexities for each data set on the smallest samples.
We chose these to highlight distinct structures at the coarsest sampling level, which is also the fastest to run, see Section~\ref{sec:ApplicationsOfTheLinearRelationship}.
We scale the perplexities linearly with the sampling sizes, following the reasoning established in Section~\ref{sec:PerplexityScalesLinearlyWithPointSetSize}.
From the four starting perplexities and the four different sampling sizes, we obtain 16 different perplexity values for each data set, see Table~\ref{tab:PerplexityValues}.

\begin{table*}[]
    \centering
    \begin{tabular}{c||r|r|r|r||r|r|r|r||r|r|r|r|}
        $\rho$ & \multicolumn{4}{|c|}{\textbf{Perplexities MNIST}} & \multicolumn{4}{|c|}{\textbf{Perplexities C.Elegans}} & \multicolumn{4}{|c|}{\textbf{Perplexities WONG}}\\
        \hline
        0.1 & 1 & 7 & 21 & 144 & 3 & 12 & 34 & 105 & 3 & 30 & 100 & 200 \\
        0.4 & 4 & 28 & 84 & 576 & 12 & 48 & 136 & 420 & 13 & 121 & 401 & 801 \\
        0.7 & 7 & 49 & 147 & 1,008 & 21 & 84 & 238 & 735 & 22 & 211 & N/A & N/A \\
        1.0 & 10 & 70 & 210 & 1,440 & 31 & 121 & 341 & 1,051 & 31 & 301 & N/A & N/A 
    \end{tabular}
    \caption{
        Perplexity values used in the embeddings of Figure~\ref{fig:LinearScalingPerplexity}.
        First row is fixed based on visual expressiveness and the other rows scale linearly based on the indicated sampling size~$\rho$.
    }
    \label{tab:PerplexityValues}
\end{table*}

Third, on each sample and the corresponding initial embeddings, we apply t-SNE to obtain a final embedding for the respective sample.
Each t-SNE run is executed with fast Fourier transform t-SNE~\cite{linderman2019fast}.
We employ 250~iterations of early exaggeration (exaggeration factor~12~\cite{belkina2019automated}, momentum~0.5~\cite{jacobs1988increased,vandermaaten2008visualizing}) and 750~iterations of gradient descent (exaggeration factor~1, momentum~0.8~\cite{jacobs1988increased,vandermaaten2008visualizing}).
The iteration numbers follow previous studies~\cite{belkina2019automated} and 
standard parameter values as used in \href{https://scikit-learn.org/stable/modules/generated/sklearn.manifold.TSNE.html}{SciKit Learn} or \href{https://opentsne.readthedocs.io/en/latest/examples/01_simple_usage/01_simple_usage.html}{OpenTSNE}.

\begin{figure*}
    \centering
    \includegraphics[width=1.\textwidth]{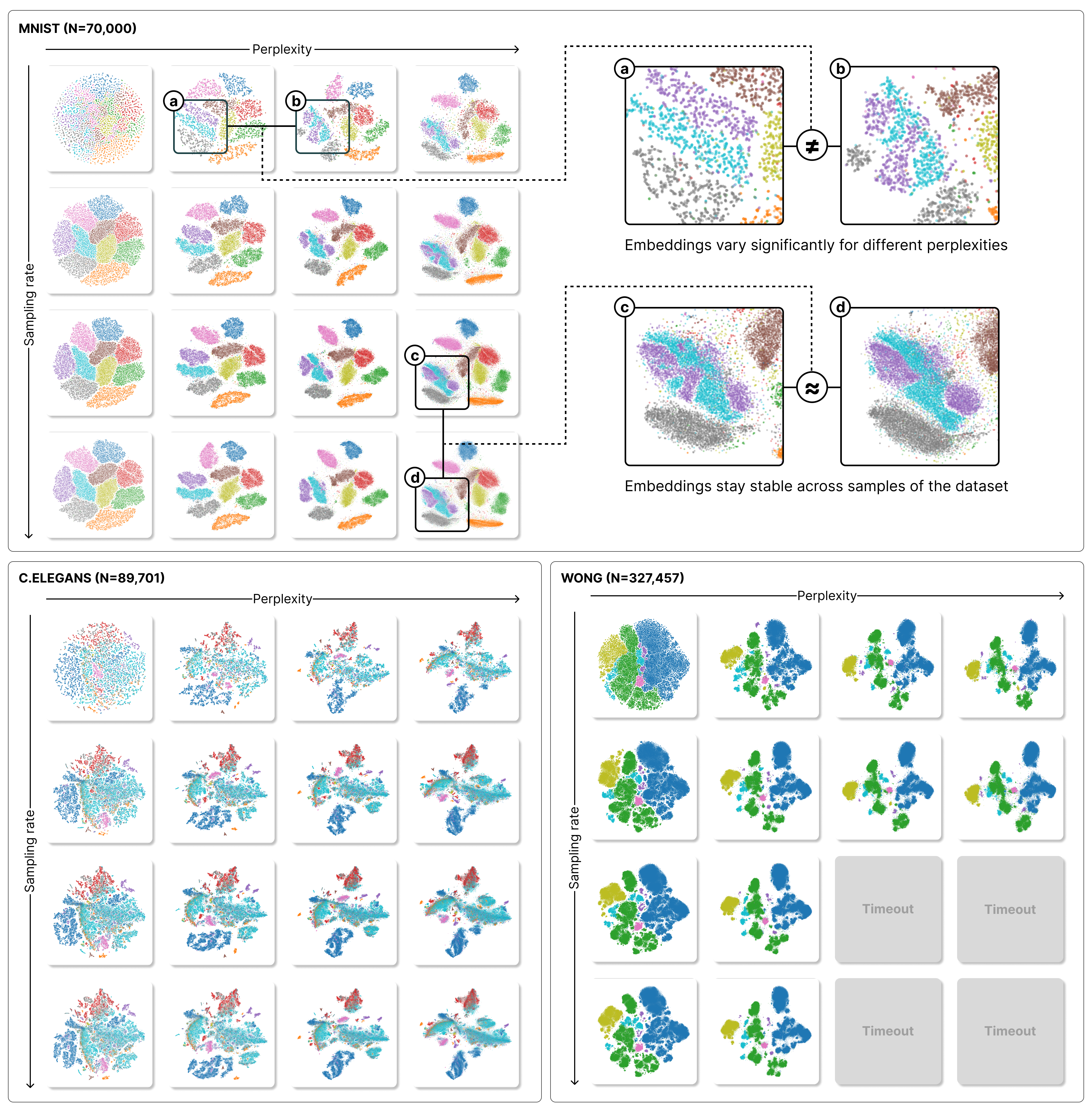}
    \caption{
        Qualitative evaluation of our results via embeddings of the MNIST, C.Elegans, and WONG data set.
        For each data set, top to bottom: We increase the number of points but keep the ratio between perplexity~$\per$ and data size~$n$ fixed.
        Also, left to right: We increase the perplexity~$\per$, but keep the data size fixed.
        The two highlight boxes on MNINST demonstrate how embeddings vary across columns in a row but remain stable across rows within a column.
    }
    \label{fig:LinearScalingPerplexity}
\end{figure*}

Finally, we present the obtained embeddings in Figure ~\ref{fig:LinearScalingPerplexity}.
In each column, from top to bottom, we increase the number of points from the data set to be embedded, while scaling the perplexity linearly.
From left to right in the rows, we increase the perplexity on the same sampling size and initial embedding.
Therefore, the three matrix arrangements of~$4\times4$ embeddings correspond to the arrangement of perplexity values in Table~\ref{tab:PerplexityValues}.

The reasoning of Section~\ref{sec:PerplexityScalesLinearlyWithPointSetSize} lets us hypothesize that the embeddings vary between rows, as the ratio of perplexity over sample size changes.
Further, we also presume that the embeddings preserve key visual characteristics across any given column since the perplexity is scaled linearly with the data set size along the columns.

Indeed, we observe both cases across all the different embeddings of the three data sets.
Consider, for instance, the highlights \textbf{a} and \textbf{b} in the first row of MNIST embeddings in Figure~\ref{fig:LinearScalingPerplexity}.
While in the highlight \textbf{a}, the purple, blue, and gray clusters form horizontal layers, an increase of perplexity causes a mixture of these three clusters in highlight \textbf{b}.
Thus, embeddings vary significantly across the rows, that is, across different perplexities, which is consistent with previous observations~\cite{wattenberg2016how}.

However, we also observe that the embeddings remain structurally stable within each column of the presented~$4\times4$ embeddings arrangements.
Consider, for example, the highlights \textbf{c} and \textbf{d} in the last column of the MNIST embeddings. 
While the highlight \textbf{ c} represents a part of an embedding of a~$\rho=0.7$ sample and highlight \textbf{d} represents a part of an embedding of the entire data set, the visual structures of the clusters remain the same, including the mixing of the blue and purple cluster.
This is due to the linear scaling of the perplexity.

Similar observations, as those made for the MNIST data set, hold for the rows and columns of the C.Elegans and WONG data sets, respectively.
Our machine had an AMD Ryzen 9, 5950X (32) with 3,400~GHz and 32~GB RAM.
Therefore, we could not compute embeddings for the two higher perplexity values and the two larger sampling sizes of the WONG data set.
These cases are marked with ``Timeout'' in Figure~\ref{fig:LinearScalingPerplexity}.
However, the stability of embeddings across columns still allows an operator to judge the embeddings present.
We will discuss such consequences of our results for visualization workflows in Section~\ref{sec:ApplicationsOfTheLinearRelationship}.



\section{Consequences for t-SNE visualization workflows}
\label{sec:ApplicationsOfTheLinearRelationship}

\begin{figure*}
    \includegraphics[width=1.\linewidth]{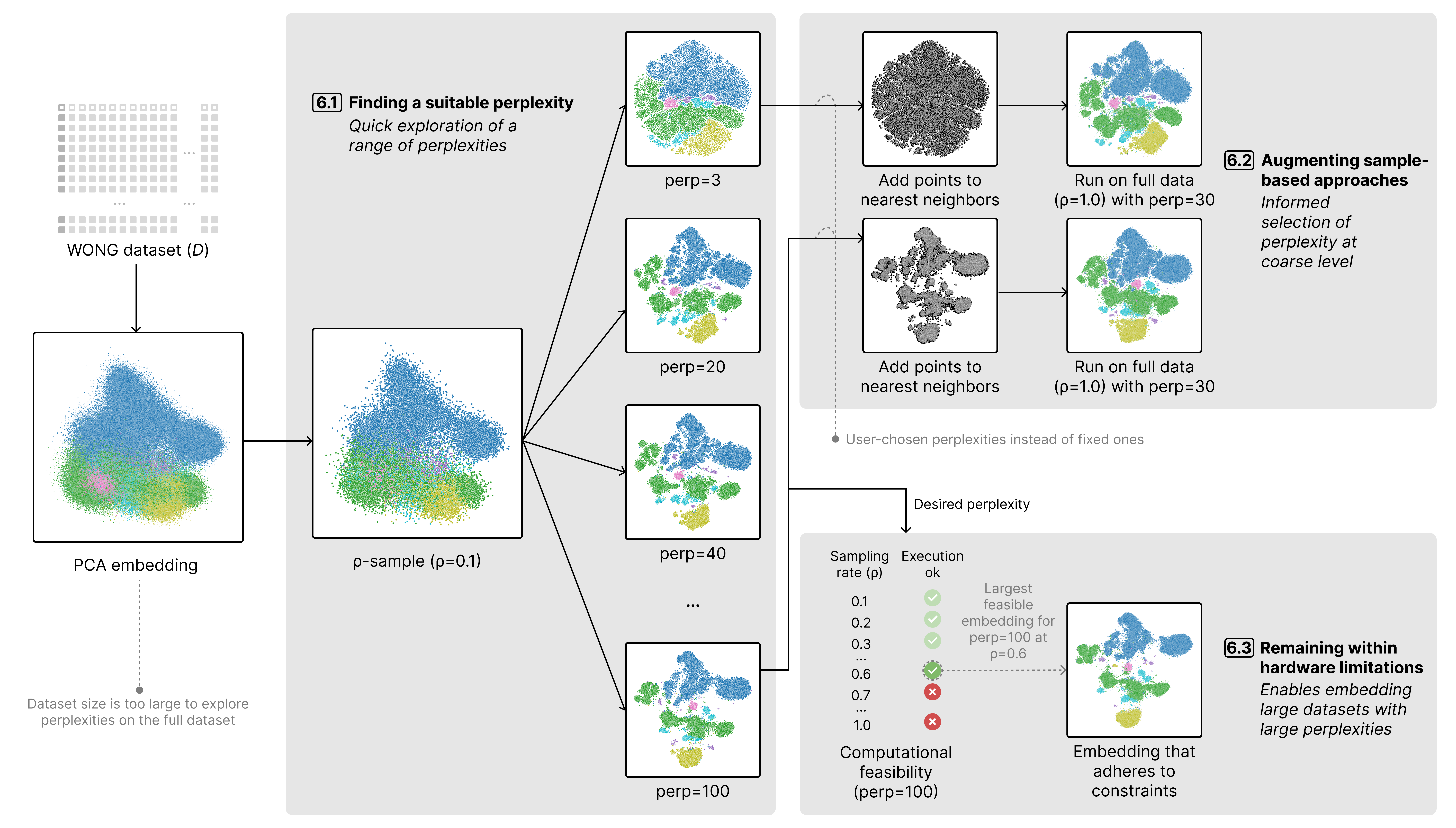}
    \caption{
        Illustration of the impact of our results on a typical t-SNE visualization workflow.
        For a data set that is too big to compute various embeddings with different perplexities, the user can quickly explore different embeddings on a small sample, Section~\ref{sec:FindingASuitablePerplexity}.
        The output of this selection process can then be used as input for a sample-based embedding approach, allowing the user to steer this approach more directly, Section~\ref{sec:AugmentingSampleBasedEmbeddingApproaches}.
        Finally, if the machine cannot compute an embedding of the full data set with the user-chosen perplexity, the user can scale sample size and perplexity to max out their hardware, thus adhering to the limitations, and still obtain a feasible embedding of an as-large-as-possible sample of the data, Section~\ref{sec:AdheringToHardwareLimitations}.
    }
    \label{fig:VisualizationWorkflow}
\end{figure*}

Consider a typical t-SNE visualization workflow, schematically shown in Figure~\ref{fig:VisualizationWorkflow}.
The user handles a data set~$D$ that they want to embed in some low-dimensional space.
To do so, they first face the challenge of choosing a suitable perplexity value, where different values might highlight varying aspects of the data.
Having chosen a perplexity, they compute the data embedding, for instance following a sample-based embedding approach~\cite{kobak2019art}.
During this process, all computations must remain feasible given the available hardware.
In this section, we will describe the consequences of our results for three aspects of this workflow: The choice of a suitable perplexity, Section~\ref{sec:FindingASuitablePerplexity}, the scheme of sample-based embeddings, Section~\ref{sec:AugmentingSampleBasedEmbeddingApproaches}, and adhering to hardware limitations, Section~\ref{sec:AdheringToHardwareLimitations}.

\subsection{Finding a suitable perplexity}
\label{sec:FindingASuitablePerplexity}

When preparing to use t-SNE for visualizing high-dimensional data, one of the first tasks is selecting a perplexity value that aligns with the data set's size and structure as well as the intended use-case of the embedding.
Especially for large data sets, fully computing several embeddings for different perplexity values quickly becomes computationally intensive.
By embedding small, representative samples of the data set with a range of perplexity values, users can identify which perplexity choices yield structures in the visualization that are meaningful for their current use-case. 
This approach not only saves computational resources but also highlights how different perplexity values emphasize varying levels of detail within the data, from broader clusters to finer-grained substructures, as observed in prior work~\cite{wattenberg2016how}.

Moreover, this sampling-based exploration allows users to develop intuition about the interplay between perplexity, data size, and the scale of patterns represented in the visualization. 
Visualizing these sample embeddings provides insights into the embedding landscape, ensuring the chosen perplexity range is neither too low---causing overly fragmented visualizations---nor too high---losing subtle local details. 
This strategy is exemplified in Figure~\ref{fig:VisualizationWorkflow}, where a small~{$\rho=0.1$} sample of the data set~$D$ is embedded with varying perplexities ranging from~{$\per=3$} to~${\per=100}$.

\subsection{User-chosen input to sample-based embedding approaches}
\label{sec:AugmentingSampleBasedEmbeddingApproaches}

A sample-based approach for t-SNE was proposed~\cite{kobak2019art}, including the following steps: 
(i) downsample a large data set to some manageable size; 
(ii) run t-SNE on the subsample using our approach to preserve global geometry; 
(iii) position all the remaining points on the resulting t-SNE plot using nearest neighbours; 
(iv) use the result as initialisation to run t-SNE on the whole data set.
While the authors prove this to be an efficient approach, they combine two different perplexity values for embedding the sampled data set in the second step:~${\per=30}$ and the heuristic~${\per = n/100}$.
Similar to other heuristics~\cite{belkina2019automated}, this restricts the user to a given, fixed perplexity which is independent of the data set size and structure.

Our findings suggest a more dynamic approach: adjusting perplexity values based on the sampling rate, thus leveraging the linear relationship between perplexity and data set size. 
This augmentation ensures the sampled embedding not only captures global geometry effectively but also aligns with the desired level of detail as dictated by user-defined preferences.
In practice, users select a perplexity value suitable for the initial subsample size, as discussed previously.
Using this user-chosen perplexity value not only preserves the robustness of the embedding but also introduces flexibility, allowing users to fine-tune the visualization's focus. 

Additional to that, computing a t-SNE embedding of the full data set after adding all data points to the sample embedding is very robust concerning the used perplexity, allowing a small value, for instance,~${\per=30}$, which is thus computationally efficient~\cite{kobak2019art}.
However, this stresses the importance of the choice of the perplexity used on the sample, for which our results provide more insight.
This is exemplified in Figure~\ref{fig:VisualizationWorkflow}, where two different, user-chosen perplexities---${\per=3}$ and ${\per=100}$---and their respective embeddings are used as input to the sample-based embedding approach.

\subsection{Adhering to hardware limitations}
\label{sec:AdheringToHardwareLimitations}

Practical limitations often arise when scaling to large data sets. 
Specifically, large perplexities for large data sets demand substantial RAM and computational resources, as the construction of the high-dimensional probabilities from Equation~(\ref{equ:HighDimensionalProbability}) grows in complexity. 
This issue, as demonstrated for the WONG data set in Figure~\ref{fig:LinearScalingPerplexity}, can impede the visualization process, especially when working with limited hardware. 
Our results offer a solution: users can scale perplexity and sample size strategically to optimize the embedding process within their hardware constraints while preserving the integrity of the visualization.

To achieve this, users first determine a suitable perplexity for a small sample as discussed above. 
This perplexity can then be linearly scaled with the sample size, up to the maximum combination of data size and perplexity the user's hardware can accommodate. 
By operating near this boundary, users maximize the data set size included in the visualization while retaining the desired embedding properties. 
This approach ensures a balance between computational feasibility and the richness of the representation, allowing users to extract as much information as possible without exceeding hardware limitations.
We exemplify this approach in Figure~\ref{fig:VisualizationWorkflow}, where we show how a column from the WONG data set from Figure~\ref{fig:LinearScalingPerplexity} can be extended to the maximum that can be run on our machine.

\section{Conclusion and Future Work}

In this paper, we uncovered a linear relationship between perplexity and data set size in t-SNE embeddings, see Proposition~\ref{pro:MainResult}.
We demonstrated that when perplexity is scaled proportionally to the size of the data set, the structural integrity of the resulting embeddings remains stable. 
This finding provides a principled framework for selecting perplexity values.
By bridging the gap between heuristic-based and systematic perplexity selection, our work contributes to more reliable and interpretable visualizations for high-dimensional data analysis.

To validate our findings, we conducted quantitative and qualitative experiments, see Section~\ref{sec:ExperimentalValidation}. 
These confirmed that embeddings produced with linearly scaled perplexities maintain their structural stability, even as sample sizes vary. 
Beyond this, we explored practical implications for t-SNE workflows, including strategies for estimating suitable perplexity values, augmenting sample-based embedding approaches, and operating effectively within hardware constraints, see Section~\ref{sec:ApplicationsOfTheLinearRelationship}.
These scenarios demonstrate how our results can be applied to enhance the usability and scalability of t-SNE in real-world scenarios.

While our method offers structural stability of key visual features, it has limitations. 
In particular, small samples may fail to represent or even obscure minor clusters in the data. 
As a result, the perplexity estimates derived from small subsets may not adequately capture these clusters. 
For use cases where small clusters are crucial, our method might fall short, and alternative approaches, such as H-SNE, which explicitly account for hierarchical structures and smaller data subsets, might be more appropriate~\cite{pezzotti2016hierarchical}. 
Recognizing these limitations underscores the importance of tailoring the choice of embedding strategies to the specific characteristics and goals of the data analysis task.

Our findings enable future research, particularly in connecting t-SNE with approaches from the domain of Partial Differential Equations (PDEs). 
Specifically, multigrid methods used in solving PDEs offer a parallel to sample-based embeddings. 
These methods progressively sample a domain, solve the coarsest representation first, and refine the solution iteratively to finer levels. 
This has a clear similarity to the different sample sizes and resulting embeddings discussed in this paper.
Exploring how multigrid techniques can be adapted to enhance t-SNE embeddings presents an interesting opportunity. 
Such an approach could enable more efficient handling of large data sets while retaining high-quality embeddings, potentially combining the strengths of sample-based methods with the refinement capabilities of multigrid frameworks.
The connections of the different layers could be established by the results in this paper.

\section*{Utilized Data Sets}
The data sets used in our experiments are available as follows: MNIST (\href{https://yann.lecun.com/exdb/mnist/}{https://yann.lecun.com/exdb/mnist/}), Cifar-100 (\href{https://www.cs.toronto.edu/~kriz/cifar.html}{https://www.cs.toronto.edu/\textasciitilde kriz/cifar.html}), C.Elegans (\href{https://github.com/Munfred/wormcells-data/releases}{https://github.com/Munfred/wormcells-data/releases}), and WONG (\href{http://flowrepository.org/id/FR-FCM-ZZTM}{http://flowrepository.org/id/FR-FCM-ZZTM}).


\printbibliography                

\end{document}